\documentclass[letterpaper, 12 pt, conference]{ieeeconf}  

\onecolumn
\usepackage{graphicx}
\usepackage{textcomp}

\IEEEoverridecommandlockouts                              
\title{\LARGE \bf
Neural Mesh: Introducing a Notion of Space and Conservation of Energy to Neural Nets
}

\author{Zoe D. Papakipos and Jacob A. Beck}%

\begin{document}

\maketitle
\thispagestyle{empty}
\pagestyle{empty}

\section{ABSTRACT}
Neural nets are based on a simplified model of the brain. In this project, we wanted to relax the simplifying assumptions of a traditional neural network by making a model that more closely emulates the low-level interactions of neurons. Like in an RNN, our model has a state that persists between time steps, so that neurons' energies stay around. However, unlike an RNN, our state consists of a 2-dimensional matrix, rather than a 1-dimensional vector, thereby introducing a concept of distance to other neurons within the state. In our model, neurons can only fire to adjacent neurons, as in the brain. Like in the brain, we only allow neurons to fire in a time step if they contain enough energy, or excitement. We also enforce a notion of conservation of energy, so that a neuron cannot excite its neighbors more than the excitement it already contained at that time step. Taken together, these two features allow signals in the form of activations to flow around in our network over time, making our neural mesh more closely model the brain, as compared to a classic neural net. Although our main goal is to design an architecture to more closely emulate the brain in the hope of having a correct internal representation of information by the time we know how to properly train a general intelligence, we did benchmark our neural mash on a specific task. We found that by increasing the runtime of the mesh, we were able to increase its accuracy without increasing the number of parameters. Moreover, we found that, with a small window size, evaluated on classifying MNIST digits, our neural net performed slightly better with a relatively small number of neurons than a feed feed-forward network with the same small number of neurons, however it performed slightly worse for large numbers of neurons. For a large window size, the reverse effect can be seen: our neural network performs worse for a relatively small number of neurons than a feed-forward network, but better when they are both given many neurons.

\section{BACKGROUND}
Neural nets were created to emulate a basic idea of how the brain works: neurons fire and excite other neurons, and how much one neuron excites another gets stronger the more they fire together. The details of the optimization algorithm have changed since Hebbian learning, given the advent of gradient descent, but the idea remains: to loosely model the brain, in the hopes that emulating one model that produces intelligence will help us get closer to reproducing actual intelligence. And neural nets do work much better than their predecessors at many tasks, but this does not mean they are actually close to producing intelligence itself.

\vspace{0.3cm}

\begin{figure}[thpb]
      \centering
      \includegraphics[scale=0.7]{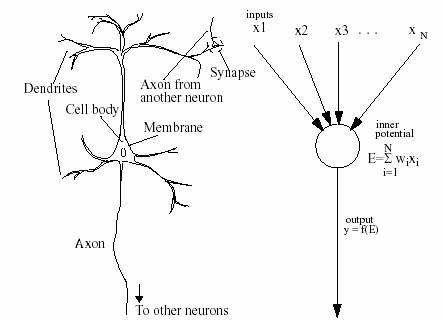}
      \caption{Classic neural nets emulate a basic model of the brain. [1]}
      \label{figurelabel}
   \end{figure}

Recurrent neural nets, or RNNs, took neural nets a step closer to how our brain works: an RNN keeps around a persistent state throughout time steps, thus keeping temporally disjoint information around. RNNs also use masking mechanisms to determine which information to keep around at each step. Both of these improvements resemble how a network of actual neurons works: energy doesn't just reset in the neurons after each time step or inference, but rather, related energy levels persist, and the state of the network keeps some information around, depending on the state at previous time steps. In these ways, RNNs are closer to a true model of the brain, but they still don’t get it quite right, because they reconfigure all parts of their entire state based upon the current input. We wanted to improve upon the RNN by imposing a notion of space and forcing the neurons to obey laws of physics that our brains are held to. If deep neural nets and RNNs were the first steps toward emulating neurons learning in the brain, then we think that our neural mesh is another step along the same path.

\vspace{0.3cm}

\begin{figure}[thpb]
      \centering
      \includegraphics[scale=0.7]{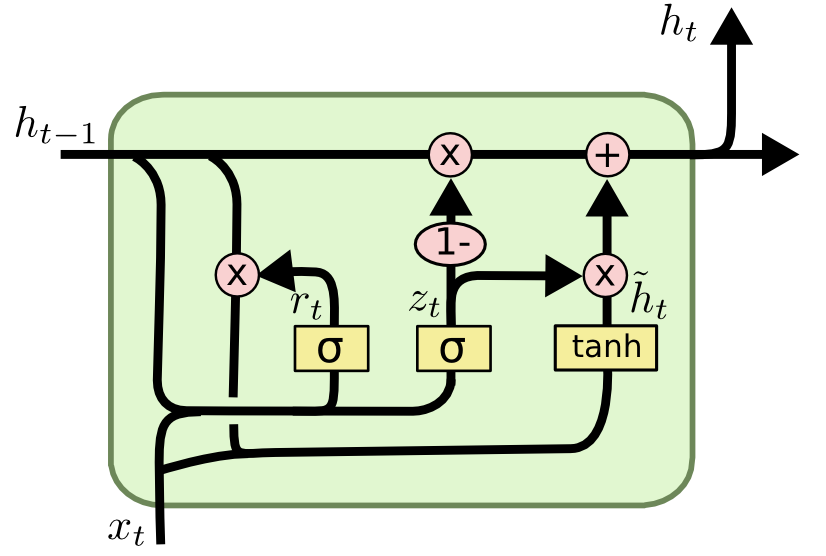}
      \caption{An RNN, like this GRU, has a state that persists between time steps, or inputs. [2]}
      \label{figurelabel}
   \end{figure}

So how do neurons actually learn? How are connections formed, and how do neurons pass energy to other neurons? Neurons in the brain exist in a network in 3D space. Each neuron can fire to up to 10,000 other nearby neurons. When a neuron gets ``excited'', ions move into it, and it ``fires'', meaning it releases some number of vesicles of neurotransmitters across each synapse to adjacent neurons. [3] Each neuron releases one of two types of neurotransmitters: inhibitory or excitory. Inhibitory neurotransmitters inhibit excitement, whereas excitory neurotransmitters increase excitement. How many vesicles are released varies changed based on the neurons involved. (The number also varies also within a given neuron-to-neuron connection, or synapse, but it’s not yet well understood in the neuroscientific community when and why a neuron releases a certain number of vesicles. It may be somewhat random, but it seems more likely the reason is something we haven't been able to identify yet). Finally, neurons have to build up excitement to a certain level before they can fire again after a refractory period. \\

In order to model these process, we make several of the assumptions already made in traditional feed-forward networks: we model a synapse by a scalar weight representing the number of vesicles being released, we model the excitory or inhibitory property of the vesicles as the sign of that weight, and we only allow the neuron to fire if it has passes a threshold (we chose 0, as with the ReLU activation).\\

So how does our neural mesh differ from a traditional neural network? In our neural mesh, we also build into our model the fact that our brains exist in a physical space with a distance metric. We do this by making the state 2-dimensional and only allowing neurons to fire into their adjacent neighbors. (Note that real brain are 3-dimensional, and traditional RNNs don't have notion of locality.) In addition, we don't feed in a new input at every time step expecting the input to be fully processed in just one time step. Instead, before we require an output, we allow time for signals to propagate through the mesh, since the signal can only move one neuron at a time in the mesh. This gives the neural mesh time to ``think''. Finally, we better model the firing phenomenon over time. That is, firing has consequences and affects the local state at the next time step, instead of simply being updated to whatever value is convenient. We do this by allowing neurons to use up their ``excitement'' in order to fire, and then force them to be depleted on the next time step. More specifically, we force that energy is conserved. Whatever energy is sent from them to their neighbors, must be removed from them (potentially with some energy lost to friction or other forces). In other words, however much excitement a neuron has in a given time step, that neuron cannot affect (excite or inhibit) other neurons any more than that amount. That is, for a neuron $i$ with activation $a_i$, neighbors in set $N_i$, and weight $w_{ij}$ from neuron $i$ to $j$:

$$\sum_{j \in N_i}{|w_{ij}a_i|} \leq a_i$$

\section{METHODS}
Our neural mesh is a network of neurons, where the location of neurons determines which neurons it is connected to. While neurons in the brain are admittedly arranged in a 3D network, we chose to model it in 2D. We simplified the geometry to a square grid, where each neuron can fire to the neurons above, below, right, and left of itself. However, the grid ``wraps around'', meaning the neurons on the top row can fire to the ones on the bottom row, and vice versa, and the neurons on the rightmost column fire to the ones on the leftmost column, and vice versa. Thus, our network can be seen as a 3D mesh - the surface of a torus, to be precise.

\vspace{0.3cm}

\begin{figure}[thpb]
      \centering
      \includegraphics[scale=0.3]{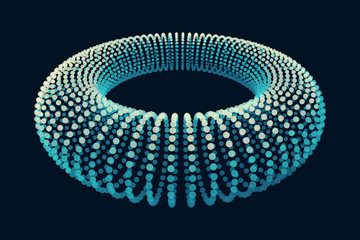}
      \caption{The effective shape of our neural mesh is a torus. [4]}
      \label{figurelabel}
   \end{figure}
   
Unless otherwise specified, we trained via SGD with a batch size of 2500 for 30 epochs with 10,000 train images and 10,000 test images and with a learning rate of 0.001. Unless otherwise specified, we used 25 neurons by 25 neurons (625 neurons total) for our mesh. In addition to the description mentioned so far, so we tried out several configuration options that are listed below. (Note, we also tried 2 others: allowing neurons to deplete all the way to the less excited state of -1 by allowing it an extra unit of energy when it fires, and normalizing the weights out of a given neuron to sum to 1 to force them to use up all of their neuron's energy when they fire. However, we found that these two configutation options performed a few and about 20 percentage points less accurately respectively and did not pursue them.) All of the properties that follow we did test, and assume that unless otherwise specified later, all of the following properties are false/not used:\\

\begin{enumerate}
\item 
Keep showing the image as a residual input at every time step, rather than just on the first one.
\item 
Max the activations with 0 at every time step (i.e. first apply ReLU to the activations).
\item
Include a bias in the feed-forward layer for the input. (The default is to not use a bias.)
\item 
Clip the activations to [-1, 1] after any addition to the state.
\end{enumerate}

\vspace{0.3cm}

\begin{figure}[thpb]
      \centering
      \includegraphics[scale=0.6]{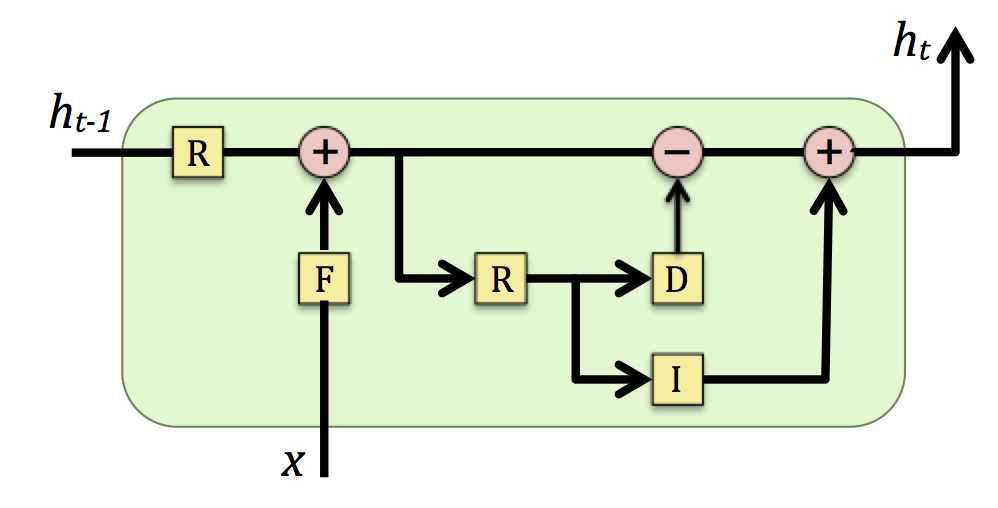}
      \caption{The neural mesh architecture (including option 2, not option 4, neutral with respect to option 1 and 3)}
      \label{figurelabel}
   \end{figure}
   
\pagebreak 

Neurons in the neural mesh fire into adjacent neurons essentially the same way as normal neural nets fire neurons: for each neuron-to-neuron connection, there is a weight that is trained via gradient descent. When a neuron $i$ fires to neuron $j$ at time $t$, neuron $j$'s activation $a_j$ for time $t + 1$ is incremented by $w_{ij} a_i$. At each time step, all the neurons update simultaneously based on other neurons firing into it.\\

To see what the architecture looks like, see figure 4. X is the input (which can be the 0 vector if the mesh is being allowed more time to process the same thing), and h is the hidden state. R refers to performing a ReLU operation, F to applying a feed-forward layer, D to a decrement operation, and I to an increment operation. A decrement operation computes the amount of energy leaving each neuron on a given timestep, and an increment operation computes the amount of energy entering a neuron. $M_u$ is the matrix that shifts an input matrix up when it is multiplied on the left of that input matrix, i.e. the identity matrix shifted one row up. $M_d$ is defined similarly except it shifts inputs down. $M_r$ is the matrix that shifts an input matrix to the right when it is multiplied on the right of that input matrix, i.e. the identity matrix shifted to the right one column. $M_l$ is defined similarly except it shifts inputs left. $W_u$ is a matrix of the weights from each neuron to its neighbor above (and $W_d$, $W_r$, $W_l$, are defined similarly but for down, right, and left respectively). $*$ represents pointwise multiplication. Given that a $W$ computes the transfer out of each neuron, the decrement matrix is computed as follows, on the current state, $s$:

$$D(s) = \frac{1}{4} |W_u * s| + \frac{1}{4} |W_d * s| + \frac{1}{4} |W_r * s| + \frac{1}{4} |W_l * s|$$

An increment operation computes the amount of energy entering each neuron, as follows:

$$I(s) = \frac{1}{4} M_u (W_u * s) + \frac{1}{4} M_d (W_d * s) + \frac{1}{4} (W_r * s) M_r + \frac{1}{4} (W_l * s) M_l$$
   
Putting this together, our neural mesh first performs ReLU on the input (if we are using option 2), resetting all negative energies to 0, then second, it adds in energies from the current input, using a fully-connected layer to get it into the right shape (bias determined by option 3), then third, it uses a ReLU to compute the neurons that contain positive activations that are available for firing. After this, if a neuron fires, it transfers some fraction of its energy to each of its neighbors, sending at most a quarter in any direction. The decrements matrix is the amount sent out of each neuron (to be subtracted) and the increment matrix is the amount that each neuron receives (to be added to the current state). These steps calculate the new state h of the neural mesh. (To get an output for our classification, we simply use a fully-connected layer with a softmax activation.)\\

Ideally, our neural mesh would be evaluated as a general AI, since our aim in coming up with this architecture was to make something that can learn like a human. But there are currently no datasets or standard benchmarks for evaluating general AIs, and we also don't have a good loss function to use even if we had such a dataset. If we were to try to evaluate our neural mesh as a general AI, we might use an evolutionary algorithm to learn a reward function for the environment that it is placed in. However, finding a way to evaluate general AI learning models is outside the scope of this paper. In this paper, we are proposing an internal structure to eventually use as part of an artificial brain. To evaluate this piece, we want to make sure it can at least learn low-level tasks. Therefore, we simply evaluate it on any of the standard machine learning datasets.\\

We decided to test our neural mesh on a classic machine learning task, but one that is not generally associated with recurrent models: pattern recognition in images (specifically, MNIST). Our hope for the neural mesh on this task is that instead of looking at an item in a sequence at each time step, and using the state to keep around information about previous items in the sequence, our model processing an image is more like a human looking at the same image over many time steps, and allowing computation to keep happening. You can kind of see our neural mesh as a kind of general form of a feed forward net: information flows into the mesh, and then flows around the mesh between nodes, being processed as in a multi-layer feed-forward neural net. This seems like a useful abstraction: rather than defining a fixed number of layers and a fixed size for each layer, we give the mesh a size and it can be run for as many time steps as necessary to complete the required computation.\\

Finally, we must note that under certain conditions, our neural network can be made equivalent to a feed-forward network of the same size, so we hope that by tuning the hyper-parameters, we can get it to perform at least as well. (For this reason, we use the feed-forward network as our benchmark.) If we run the neural mesh without repeatedly showing the input (no option 1), use ReLU or clipping activations (options 2 or 4), and include bias on the input (option 3), then our neural mesh can exactly emulate a one-layer feed-forward neural, as long as it runs for at least 2 time steps. It will be equivalent in the case that it learns to set all the internal weights to 0. In this case, it just takes in the input, applies the feed-forward transformation on the input and adds it to the state, leaves the state alone, then applies an activation at the beginning of the next time step. (The residual must not be used so that the state is left as is on this next time step.)\\

\section{RESULTS}
We evaluated our neural mesh against a one-layer feed-forward neural net on a classic image recognition problem: classification of MNIST handwritten digits. We found that, with a small window size, our neural net performed slightly better with a relatively small number of neurons than a feed feed-forward network with the same small number of neurons, however it performed slightly worse for large numbers of neurons. For a large window size, we found the reverse. This can be seen in figure 5 and 6 respectively.\\

\begin{figure}[!htbp]
      \centering
      \includegraphics[scale=0.75]{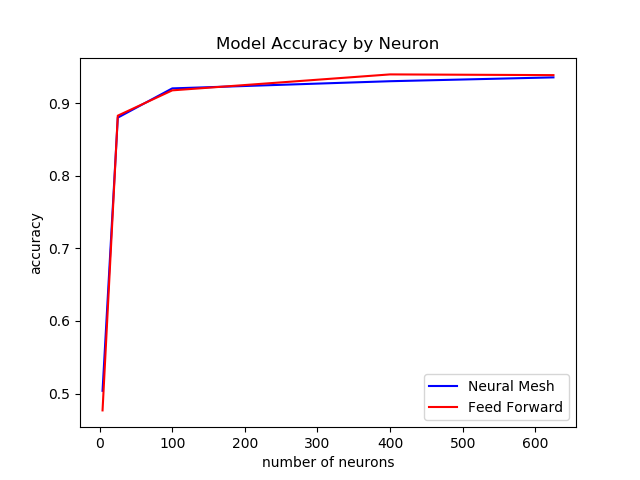}
      \caption{Our neural mesh with a smaller window size of 25}
      \label{figurelabel}
   \end{figure}
   
\begin{figure}[!htbp]
      \centering
      \includegraphics[scale=0.75]{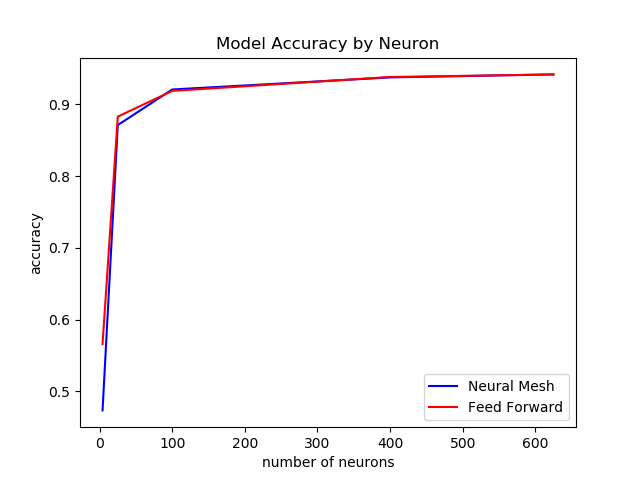}
      \caption{Our neural mesh with a larger window size of 100}
      \label{figurelabel}
   \end{figure}

\vspace{4.0cm}
In order to see that our neural mesh performs better with a large number of neurons and a large window size, see figure 7, which modifies the scale of figure 6 to zoom in on the end of the plot.

\begin{figure}[!htbp]
      \centering
      \includegraphics[scale=0.75]{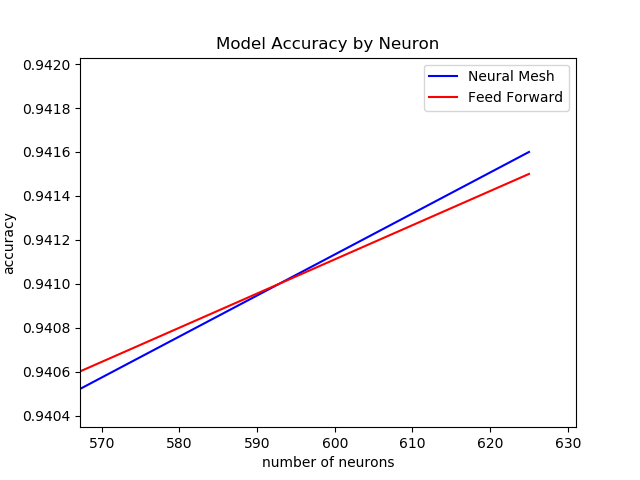}
      \caption{Our neural mesh with a larger window size of 100, zoomed in}
      \label{figurelabel}
   \end{figure}

In general, we also found that clipping (option 4) performed very similarly to the feed-forward network, however, we did not explore clipping much since it didn’t get better with windows size (see figure 8). (We assume that using a ReLU at the beginning to reset the energies - option 2 - would have a similar trend, since it is a very similar activation function, but did not test it, other than to note that its performance at window size 10 was very similar to that of the clipping and to the feed-forward network.)\\

For these reasons, in general, we focused our model on not using these activation functions at the beginning to ``reset'' the state (option 2 or 4). This means that the only time we use an activation function, is to calculate our increment, I, and decrement, D, which are then added to the state. In this way, the neural mesh is forced to use the firing mechanism to make use of the activation function. Moreover, as you can see in figure 9, our model (without option 2 or 4) gets better with longer window sizes. This suggests that our firing mechanism, used over multiple time steps, can replace a traditional activation function, and that it leverage these activations more and more with a larger window size. \\

\vspace{10.0cm}

\begin{figure}[!htbp]
      \centering
      \includegraphics[scale=0.75]{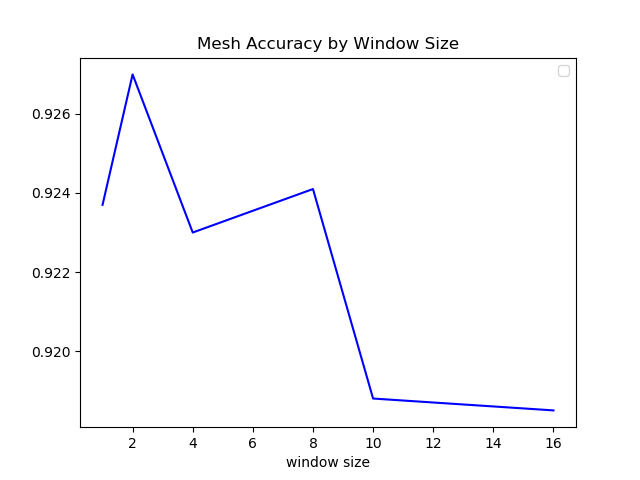}
      \caption{Increasing the window size doesn't improve the model when we clip the activations}
      \label{figurelabel}
   \end{figure}


\begin{figure}[!htbp]
      \centering
      \includegraphics[scale=0.75]{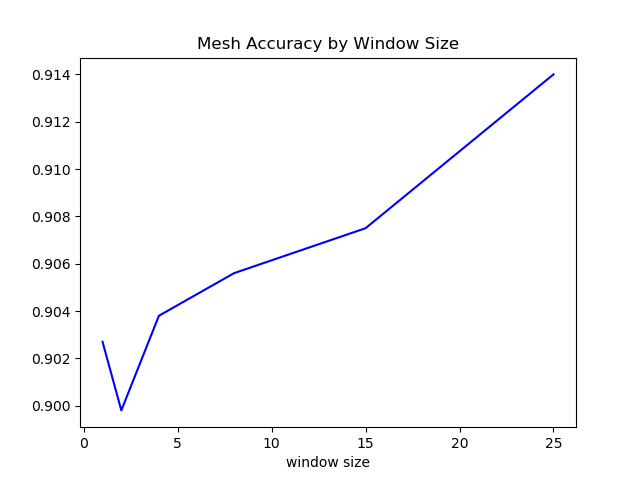}
      \caption{Increasing the window size does improve the model when we don't clip the activations}
      \label{figurelabel}
   \end{figure}

In addition to the plots above, we wanted to see how the energy moves around in the neural mesh over time to watch it ``think''. In order to do this, we created a 35 by 35 neural mesh (with 1,225 neurons), and ran it for 25 time steps on each of three randomly selected digits. We visualized this process in figures 10 and 11. Black represents a value of 0, yellow represents a value of 1, and pink is in between. In order to map the activations into [0,1], we used sigmoid for figure 10 and simply clipped the values in figure 11. (Therefore, figure 11 shows all negative numbers as black, all large positive numbers as yellow, and fractions as a shade of pink.) Note that all of the neural nets started with an initial all 0 state, which has been either mapped to pink or black. Figures 12 and 13 were produced in the same way, except they had a window size of only 10 before they output an answer.\\

We can see in these images that the total energy of the neural mesh decreases over time. This occurs because energy used by a negative weight will take positive energy from the neuron comes out of but decrease the energy of the neuron it is transferred to. Moreover, since we do not have a residual image or any resetting of energy in our default configuration, new energy cannot be introduced. This enforces that the computation at the end must come from activation initially introduced at the beginning.\\

\vspace{0.3cm}

\begin{figure}[thpb]
      \centering
      \includegraphics[scale=0.51]{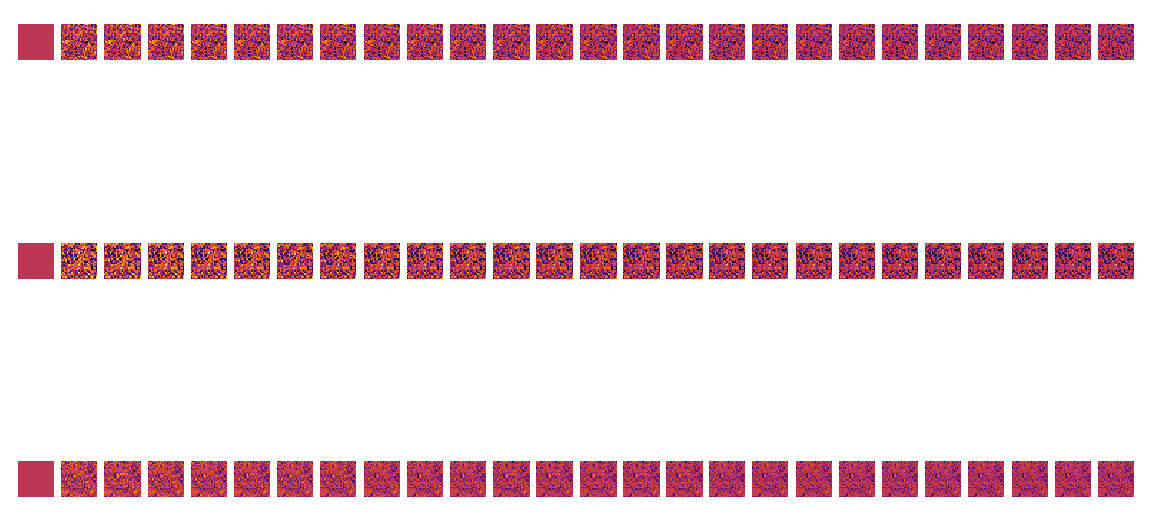}
      \caption{Activation energies over 25 time steps for 35x35 neural mesh, normalized by sigmoid}
   \end{figure}

\begin{figure}[thpb]
      \centering
      \includegraphics[scale=0.51]{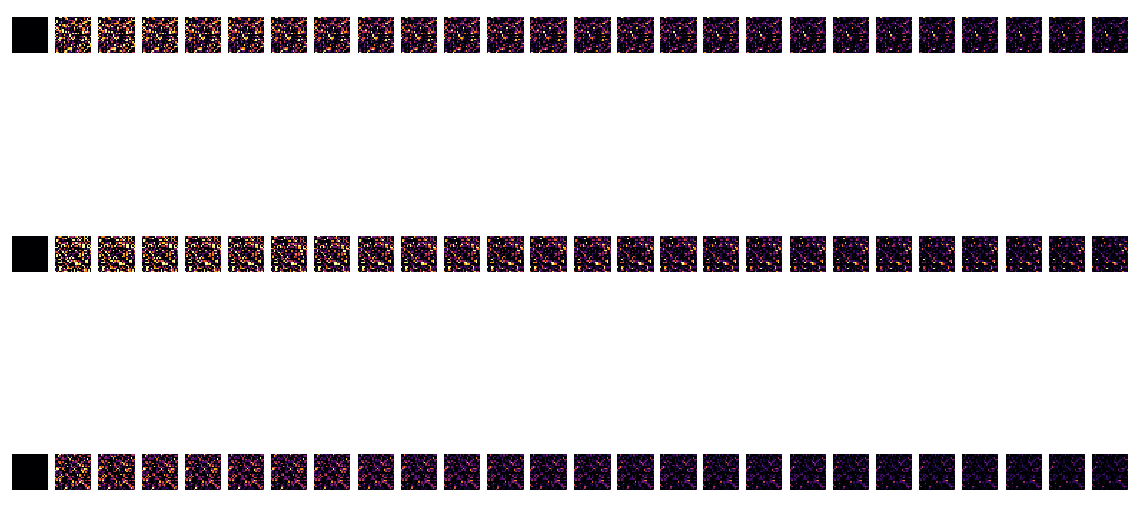}
      \caption{Activation energies over 25 time steps for 35x35 neural mesh, clipped to [0,1]}
   \end{figure}

\begin{figure}[thpb]
      \centering
      \includegraphics[scale=0.51]{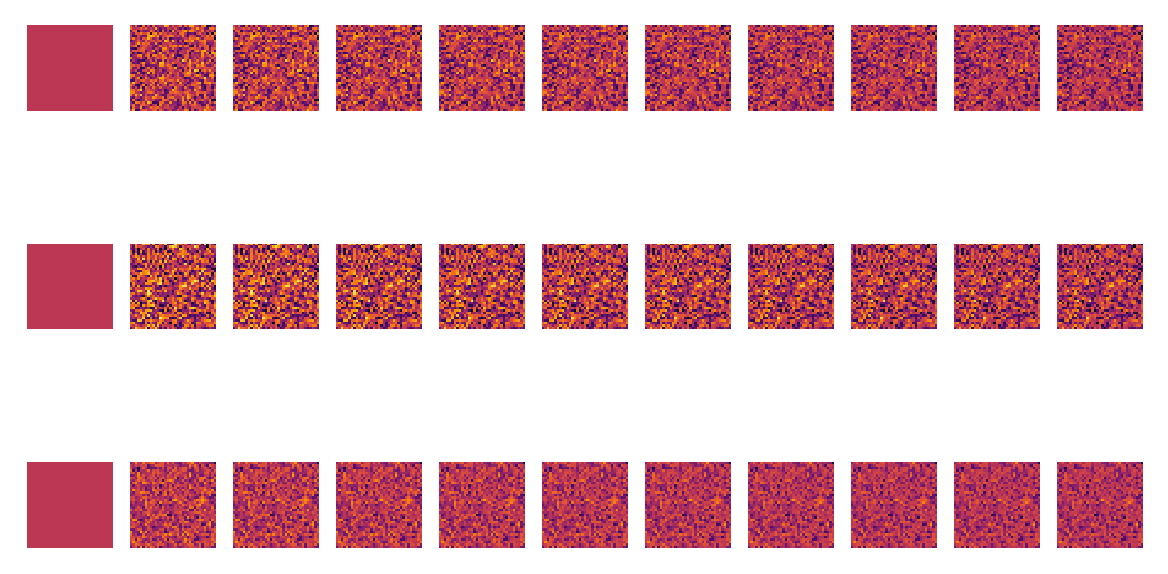}
      \caption{Activation energies over 10 time steps for 35x35 neural mesh, normalized by sigmoid}
   \end{figure}

\begin{figure}[thpb]
      \centering
      \includegraphics[scale=0.51]{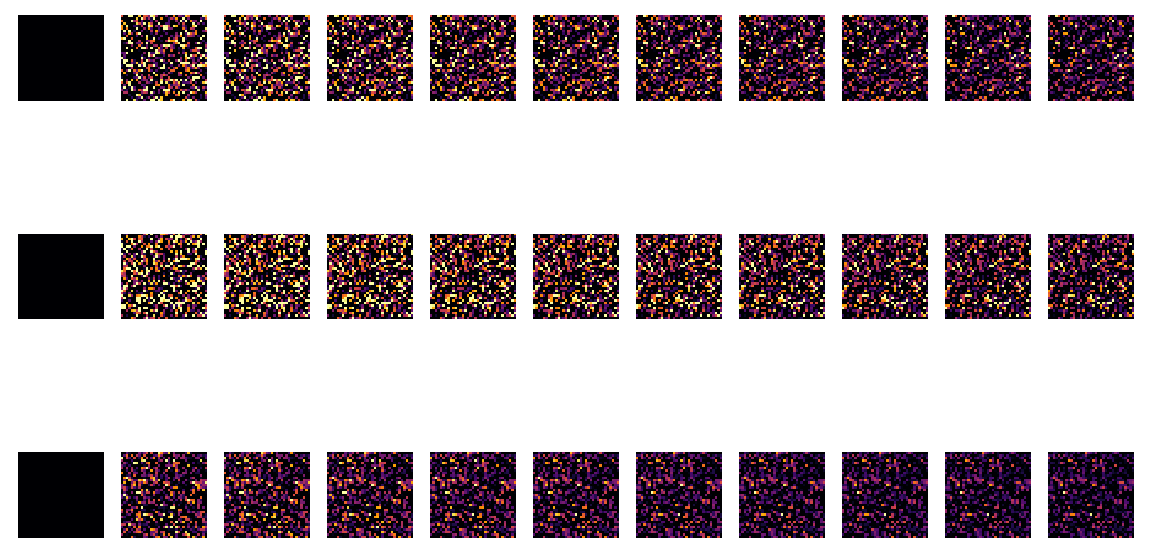}
      \caption{Activation energies over 10 time steps for 35x35 neural mesh, clipped to [0,1]}
   \end{figure}

\pagebreak

\section{CONCLUSION}
We have demonstrated that our neural mesh performs very similarly to a feed-forward network, and in certain conditions, performs a bit better given the same space complexity. Thus, our model might be useful for certain specific intelligence tasks. However, our main hope is that by designing an architecture to more closely emulate the brain, we have made something that might be useful as a general intelligence in the future. As of now, this is impossible to benchmark: we do not know the correct loss function (or reward function) to learn our parameters. However, we hope that when we do figure this out, we will be ready with the correct internal representation. Then at that point, we can try to loosen the similarities to the human brain - the one example of a general AI that we do have - and empirically see what is necessary\\

\section{FUTURE RESEARCH}
First, in order to easily configure our networks to use a residual image or not use a residual image (option 1), we simply changed the image vector to the 0 vector for all the remaining inputs in the case that we did not want a residual. However, this means that the bias was still shown at every time step. Therefore, our tests with bias actually impacted every input, even in the case where there should not have been any residual input, thus introducing (or removing) new energy to the system at every step. It may be worth seeing whether a bias on the input only on the very first step would be helpful. In this way, we can have bias but still force the end result to be processed based on the activations from the first time step. (However, in initial trials, over the course of three runs, there appears to be almost no difference in the mean performance - 93.30 without bias on the input and 93.24 with bias on just the first input.)\\

Second, we fully acknowledge that our training process is stochastic, and so our plots should have included a variance. We did create multiple plots in the course of our work that verify the trends that we present, however, due to hardware constraints, creating more copies of our benchmark plots did not seem worth the investment of time. This could be explored further, however, our main goal is not to show that our mesh is marginally better or worse with certain numbers of neurons, but rather to show that we can perform specific AI tasks with our mesh (perhaps better in certain cases) and that our mesh is closer to how the human brain works.\\

Third, in order to maintain conservation of energy, we enforce that a neuron cannot send more than $\frac{1}{4}$ of its energy in a given direction. We also tried, without promising results, to normalize the weights out of a given neuron to sum to 1 instead. There may be more ways to enforce this conservation of energy that could be explored. Additionally, since we do not fully understand how the release of vesicles with neurotransmitters works, we could try removing the scaling by $\frac{1}{4}$ or the absolute values in the decrement equation, to play around with which features of the conservation of energy if any, are important\\

Fourth, we found that reshaping the neural mesh without varying the number of neurons (e.g. 1200 x 1 versus 30 x 40) did not significantly change the behavior. This means that moving from 2-dimensional to 1-dimensional space, although it limits the paths that activation can flow, did not significantly reduce performance. This was surprising and could be explored further.\\

Finally, although we showed that our neural mesh can achieve marginally better performance under the condition that it has a lot of neurons and a large window size, we are unsure if the neural mesh will continue to get better with a larger window size or if we hit a plateau in performance. We were unable to extend the window size much more due to hardware constraints, but this would be an interesting avenue to pursue. Perhaps with longer windows we could mimic deeper neural networks. If we could continue to make gains indefinitely by just running the mesh for longer, than we would have a way to gain ever better performance without having to increase the number of parameters or space complexity.\\

\pagebreak
\section{REFERENCES}
\begin{itemize}
    \item [1.]
    Hemming, Cecilia.  Using Neural Network to Retrieve Lexical Data. Swedish National Graduate School of Language Technology, www.hemming.se/gslt/LingRes/NeuralNetworks.htm.\\
    \item [2.]
    Olah, Christopher. “Understanding LSTM Networks.” Understanding LSTM Networks, Colah Blog, 27 Aug. 2015, colah.github.io/posts/2015-08-Understanding-LSTMs/.\\
    \item [3.]
    “Signal Propagation: The Movement of Signals between Neurons.” Khan Academy, www.khanacademy.org/test-prep/mcat/organ-systems/neural-synapses/a/signal-propagation-the-movement-of-signals-between-neurons.\\
    \item [4.]
    \textcircled{c} Adobe Stock\\
\end{itemize}




\end{document}